\documentclass{article}

\usepackage{microtype}
\usepackage{graphicx}
\usepackage{subfigure}
\usepackage{booktabs} 

\usepackage{xr-hyper}

\usepackage{hyperref}

\usepackage[accepted]{icml2021}

\usepackage{amsmath,amssymb}
\usepackage{amsthm}
\usepackage{bm}
\usepackage{mathtools}

\newtheorem{theorem}{Theorem}[section]

\newtheorem{assumption}{Assumption}[section]

\newcommand{\E}[2][]{\mathbb{E}_{#1}\left[#2\right]}

\newcommand{\Exp}[1]{\exp\left(#1\right)}
\newcommand{\I}[1]{\mathbb{I}\left[#1\right]}
\newcommand{\PD}[2]{\frac{\partial{#1}}{\partial{#2}}}
\newcommand{\V}[1]{\bm{#1}}
\newcommand{\abs}[1]{\left\lvert#1\right\rvert}

\DeclareMathOperator*{\argmax}{arg\,max}

\newcommand{\Gam}[1]{Gam\left(#1\right)}
\newcommand{\Real}{\mathbb{R}}
\newcommand{\Integer}{\mathbb{Z}}

\newcommand{\HH}[1]{\mathcal{H}_{#1}}
\newcommand{\Time}{\mathcal{T}}
\newcommand{\target}{\mathrm{x}}
\newcommand{\context}{\mathcal{C}}
\newcommand{\merged}{\mathcal{M}}
\newcommand{\St}{S_{\target}}
\newcommand{\Sc}{S_{\context}}
\newcommand{\Sm}{S_{\merged}}
\newcommand{\Ht}[1]{\HH{#1}^{\target}}
\newcommand{\Hm}[1]{\HH{#1}^{\merged}}
\newcommand{\Score}{s}

\newcommand{\Len}{LEN}
\newcommand{\Rand}{RND}
\newcommand{\True}{GT}
\newcommand{\PPOD}{PPOD}
\newcommand{\CPPOD}{CPPOD}

\icmltitlerunning{Event Outlier Detection in Continuous Time}

\begin{document}
\twocolumn[
\icmltitle{Event Outlier Detection in Continuous Time}




\begin{icmlauthorlist}
\icmlauthor{Siqi Liu}{pitt,bai}
\icmlauthor{Milos Hauskrecht}{pitt}
\end{icmlauthorlist}

\icmlaffiliation{pitt}{Department of Computer Science, University of Pittsburgh, Pittsburgh, PA, USA}
\icmlaffiliation{bai}{Borealis AI, Vancouver, BC, Canada}

\icmlcorrespondingauthor{Siqi Liu}{siqiliu@cs.pitt.edu}
\icmlcorrespondingauthor{Milos Hauskrecht}{milos@pitt.edu}

\icmlkeywords{Machine Learning, ICML}

\vskip 0.3in
]



\printAffiliationsAndNotice{}  

\begin{abstract}
    Continuous-time event sequences represent discrete events occurring in continuous time. Such sequences arise frequently in real-life. Usually we expect the sequences to follow some regular pattern over time. However, sometimes these patterns may be interrupted by unexpected absence or occurrences of events. Identification of these unexpected cases can be very important as they may point to abnormal situations that need human attention. In this work, we study and develop methods for detecting outliers in continuous-time event sequences, including unexpected absence and unexpected occurrences of events. Since the patterns that event sequences tend to follow may change in different contexts, we develop outlier detection methods based on point processes that can take context information into account. Our methods are based on Bayesian decision theory and hypothesis testing with theoretical guarantees. To test the performance of the methods, we conduct experiments on both synthetic data and real-world clinical data and show the effectiveness of the proposed methods.
\end{abstract}

\section{Introduction}

\noindent Continuous-time event sequences are defined by occurrences of various types of events in time. Event sequences may represent many real-world processes and observations including, e.g., arrival of packets to servers in network systems, or administration of drugs to patients. Continuous-time event sequences are typically modeled as point processes \citep{daley_introduction_2003}. While many point-process models have been developed in the literature in recent years, most works build and test models on event prediction tasks. The focus of this work is on outlier detection in event sequences that aims to identify unusual occurrences or absence of events in event sequences in real-time. Outlier detection is the basis of many critical real-world applications, such as fraud detection \citep{Fawcett:1997}, network intrusion surveillance \citep{garcia:2009}, disease outbreak detection \citep{Wong:2003:ICML}, and medical error detection \citep{Hauskrecht:2013,hauskrecht2016outlier}.

\begin{figure}[htb]
    \centering
    \includegraphics[width=0.8\linewidth]{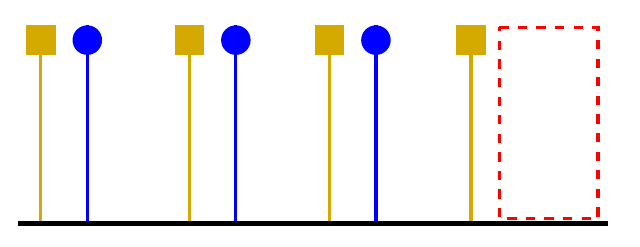}
    \includegraphics[width=0.8\linewidth]{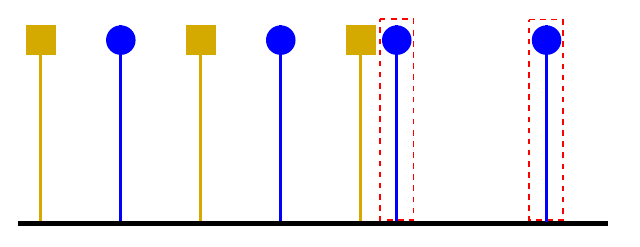}
    \caption{Examples of omission (top) and commission (bottom) outliers. Each stem and rectangle / circle represent one event on the horizontal time line. Different colors and shapes represent different types of events. Red dashed boxes mark the outliers. Top: A blue-circle event is expected but has not been observed. Bottom: The second to last blue-circle event happened too early. The last blue-circle event was not expected given the occurrence pattern.}\label{fig:examples}
\end{figure}

Two types of outliers may arise in continuous-time event sequences. First, given the history and the recent absence of the events, the event may be overdue. We refer to these as \emph{omission outliers}. Second, given the history, the event that has just arrived is unexpected: it either arrived too early or was not expected to occur at all given the context from the history. We refer to these as \emph{commission outliers}. Figure~\ref{fig:examples} shows some examples of these two types of outliers.

Both types of outliers are often related to problems of practical importance. Take for example, a person suffering from a disease and taking specific medications on a regular schedule to treat the disease. Given the history and the current time, we may infer that the person has not taken the medication yet, and the medication is overdue (omission). The detection of the overdue medication can be then used to generate a reminder alert. For commission, consider a patient who takes a medication too early compared to the normal schedule. The detection of this event or its prevention is extremely important and may prevent adverse situations like high concentration of the drug and its possible toxic effects. Similar situations may happen when one receives a medication that is unrelated to his/her condition (i.e., context). The occurrence of this event may indicate a medical error, and again its timely detection is extremely important. 

Despite the importance of these types of outlier detection problems, to the best of our knowledge, neither of them have been studied in the literature. The key contributions of this paper are the formulation of these outlier detection problems and the development of their algorithmic solutions. Specifically, we develop semi-supervised outlier detection methods \citep{chandola_anomaly_2009}, where data generated by the normal point process are available, based on Bayesian decision theory and hypothesis testing that can leverage a variety of point-process models. We show the effectiveness of the methods on both synthetic and real-word clinical event sequences. Since event occurrences often depend on other related events, we adapt an existing model, the continuous-time LSTM \citep{mei_neural_2017}, to account for these events and their history as context. However, we stress that the proposed outlier detection methods can be combined with any type of point-process model.

\section{Related Work}\label{sec:related:work}

\paragraph{Point Processes}

Point processes are probabilistic models for discrete points in continuous domains \citep{daley_introduction_2003}. They have been widely used to model continuous-time event sequences. For a temporal point process, the conditional intensity function (CIF) characterizes the process. Some researchers have proposed to use Gaussian processes to model the CIF \citep{adams_tractable_2009,rao_gaussian_2011,lloyd_variational_2015,lloyd_latent_2016,ding_bayesian_2018,liu_nonparametric_2019}, while the others have developed flexible models based on Hawkes processes \citep{hawkes_spectra_1971,zhou_learning_2013,wang_isotonic_2016,xu_learning_2016,lee_hawkes_2016,apostolopoulou_mutually_2019}, where the CIF depends on the history explicitly. Recently, neural-network-based point-process models \citep{du_recurrent_2016,mei_neural_2017,xiao_wasserstein_2017,li_learning_2018,jia_neural_2020} have shown promising results in different settings. Despite the large amount of works on point processes for event sequences, none of them have studied the problem of defining and detecting outliers in continuous time.

\paragraph{Point Processes for Noisy Data}

Recently, researchers have started developing methods to deal with noisy data, such as incomplete or missing data \citep{xu_learning_2017,shelton_hawkes_2018,mei_imputing_2019} and desynchronized data \citep{trouleau_learning_2019}. They assume the data has been corrupted by some source (e.g., censoring or noise), and the goal is to recover the original data or to learn a model nonetheless. Although the omission outliers we consider is related to missing data, our goal is to detect these outliers as accurately as possible, i.e., to distinguish them from normal data, instead of data recovery. Moreover, our method is assumed to be executed in an online manner, so we must decide whether there is an outlier based \emph{only} on the history. We do not have access to the data in the future as in the missing data setting. Finally, detecting commission outliers is not related to any of these works.

\paragraph{Outlier Detection}

In general, outlier (or anomaly) detection \citep{chandola_anomaly_2009,aggarwal:2013:book} aims to identify data instances that are unusual when compared to other instances in data. When outlier detection is applied on a subset of dimensions given the rest as context, it is referred to as contextual (or conditional) outlier detection \citep{Hauskrecht:2007,song_conditional_2007,chandola_anomaly_2009}. Our goal in this work is to detect unusual absence and occurrences of events in event sequences that depend on other types of events defining the context. Unlike outlier detection in traditional time series \citep{gupta_outlier_2013}, where time is treated as merely the index of the observed values, for continuous-time event sequences, the time of each event is essentially a random variable defined by the generative process, and the goal is to detect unusual behaviors in time given the context.

\paragraph{Semi-Supervised Detection}

Semi-supervised outlier (or anomaly) detection is a term coined in \citep{chandola_anomaly_2009}. It means that normal data (without any outliers or other corruptions) are available for training the model to detect outliers on test data later. It is a widely-adopted approach in both outlier detection \citep{fiore_network_2013, akcay_ganomaly:_2019} and other related tasks, such as intrusion detection \citep{xu_intrusion_2010} and missing data imputation \citep{mei_imputing_2019}, although in some cases, the assumption is not clearly stated. Our work also adopts this approach, but we note that event outlier detection in continuous time has not been studied in any of the previous works.

\section{Method}

\subsection{Problem Formulation}

First, we formally define the problem of contextual outlier detection in continuous-time event sequences. An event sequence can be formulated as $\St = \{(t_i: t_i \in \Time\}_{i=1}^{N_{\target}}$, i.e., a sequence of timestamps $t_i$ of the events, where $t_i$ is the time of the $i$-th event, $N_{\target}$ is the total number of the events in the sequence, and $\Time \subseteq \Real$ is the domain of time. We call $\St$ the \emph{target sequence} and the events the \emph{target events}, because they are the targets in which we aim to detect outliers. Meanwhile, we may observe contextual information along with $\St$. We assume the contextual information can be either represented as or converted to discrete events. We denote these events as $\Sc = \{(t_i, u_i): t_i \in \Time, u_i \in \context \}_{i=1}^{N_{\context}}$, where $t_i$ and $u_i$ are the time and type (mark) of the $i$-th event, $N_{\context}$ is the total number of the events, and $\context \subseteq \Integer$ is the finite set of distinct marks for different types of events. We call $\Sc$ the \emph{context sequence} and the events the \emph{context events}.

We stress that $\St$ and $\Sc$ share the same time domain $\Time$ and, therefore, we can combine them into a single sequence $\Sm = \{(t_i, u_i): (t_i,u_i)\in \Sc \text{ or } t_i \in \St, u_i = \target\}$, where a new type $\target \notin \context$ is assigned to all the events in $\St$. For detecting outliers, we only rely on information in the \emph{past} from both $\St$ and $\Sc$. We denote the combined history of $\St$ and $\Sc$ up to time $t$ as $\Hm{t} = \{(t_i, u_i): (t_i, u_i) \in \Sm, t_i < t\}$. Meanwhile, $\Ht{t} = \{t_i: t_i \in \St, t_i < t\}$ is the history of the target sequence $\St$ only without any contextual information. Now, we are ready to define two types of outlier detection problems we wish to solve.

\paragraph{Commission Outlier Detection}

Given an observed target event at time $t_n$ and the combined history $\Hm{t_n}$ up to time $t_n$, the goal is to assign a label $y_c(t_n) \in \{0, 1\}$ to $t_n$ indicating whether it is a commission outlier. Notice that $y_c(t)$ is only defined if $t$ is the time of a target event. In this work, instead of hard labels, we calculate a commission outlier score $\Score_c(t_n)$ for $t_n$ to indicate how likely it is to be a commission outlier.

\paragraph{Omission Outlier Detection}

Define a blank interval $B \subseteq \Time$ as an interval in which there is no event of type $\target$ (target event). Given a blank interval $B = (t_b, t_e)$ and the combined history $\Hm{t_b}$ up to time $t_b$, the goal is to assign a label $y_o(B) \in \{0, 1\}$ to $B$ indicating whether there are \emph{any} omission outliers in $B$. Notice that $y_o(B)$ is only defined when $B$ is an interval with no events from $\St$. In this work, instead of hard labels, we calculate an omission outlier score $\Score_o(B)$ for $B$ to indicate how likely it is to contain any omission outliers. 

\subsection{Probabilistic Models}\label{sec:model}

We develop algorithms for detecting outliers in continuous-time event sequences based on probabilistic models, specifically (temporal) point processes. Point processes are probabilistic models for discrete points in continuous domains. For an event sequence, the points are the events, and the domain is the time $\Time$. In this case, the models are also called \emph{temporal} point processes. A temporal point process can be defined as a counting process $N(\cdot)$ on $\Time$, where $N(\tau)$ is the number of points in the interval $\tau \subseteq \Time$.

For a temporal point process, the conditional intensity function (CIF), $\lambda(t)$, characterizes the probability of observing an event in an infinitesimal time interval $[t,t+dt) \subseteq \Time$ given the history up to time $t$, i.e.
$
    \lambda(t) dt = p(N([t,t+dt)) = 1|\HH{t}).
$
For our problem, we only model the \emph{target} events using a point process, while the history $\HH{t} = \Hm{t}$ contains both the target events and the context events. Because we always condition on $\HH{t}$, we omit $\HH{t}$ in notations for the rest of the paper.

For a sequence of target events $\St = \{t_i: t_i \in \Time\}_{i=1}^{N_{\target}}$ generated from the point process with CIF $\lambda(t)$, the probability density is
\begin{equation}\label{eq:pdf}
    p(\St) = \prod_{i=1}^{N_{\target}} \lambda(t_i) \Exp{-\int_{\Time}\lambda(s)ds}.
\end{equation}

In this work, we focus on semi-supervised outlier detection \citep{chandola_anomaly_2009}, meaning that we can obtain data generated from the normal point process (without outliers), from which we may train a model for detecting outliers on unseen data later. It has been widely adopted in previous works (see Section~\ref{sec:related:work}).
\begin{assumption}\label{assum:norm}
    Data generated by the normal point process are available.
\end{assumption}
We note that by definition, the majority of the data are normal rather than outliers, so obtaining normal data is usually much easier than finding and labeling outliers. Meanwhile, robust learning of the model in presence of outliers is an important but orthogonal direction for future work, since we can combine robust learning with our proposed outlier detection methods to eliminate Assumption~\ref{assum:norm}.

In general, the type of model to use should be able to represent the dependencies between the target events and the context events. For a point-process model, it means that the CIF, instead of only depending on the target events, $\lambda(t) = f(\Ht{t})$, should also depend on the context events, $\lambda(t) = f(\Hm{t})$, where $f(\cdot)$ denotes the mapping represented by the model.

In this work, we use a model adapted from the continuous-time LSTM \citep{mei_neural_2017} due to its good performance in previous studies \citep{mei_neural_2017,jia_neural_2020}. See Appendix~\ref{sec:ctlstm} for details of the model. However, we stress that if there is a better model, we can simply plug it in to get better performance.

\subsection{Detecting Commission Outliers}\label{sec:commiss}

Without knowing the true process that generates commission outliers, we make the following assumption when developing our method and later generalize it by relaxing the assumption of constant rate.
\begin{assumption}[Independent and Constant Commission]\label{assum:co}
    Commission outliers are generated by a process with a constant CIF independently from the normal point process.
\end{assumption}

Suppose we are given a target event $t_n$ and the history up to time $t_n$. Define a random variable $Z_n$, such that $Z_n=1$ if $t_n$ is a commission outlier, and $Z_n = 0$ otherwise. We are interested in calculating $p(Z_n=1|t_n)$.

By Assumption~\ref{assum:co} (Independent), the generative processes for the normal points and outliers can be viewed together as a marked point process. For each event $t_n$, there is a hidden mark $Z_n$ indicating whether it is an outlier. The overall CIF is $\lambda_g(t) = \lambda_1(t) + \lambda_0(t)$, where $\lambda_0(t)$ is the CIF for the normal point process, and $\lambda_1(t)$ for the outlier process. Suppose we are at time $t_c$. The density of having the next event at $t_n$ marginally and jointly with mark $Z_n=1$ are respectively
\begin{align}
    p(t_n) &= \lambda_g(t_n) \Exp{-\int_{t_c}^{t_n}\lambda_g(t)dt}, \\ 
    p(Z_n=1, t_n) &= \lambda_1(t_n) \Exp{-\int_{t_c}^{t_n}\lambda_g(t)dt}.
\end{align}
Then we can derive the conditional distribution of the hidden mark
\begin{equation}\label{eq:commiss:post}
p(Z_n=1|t_n) = \frac{\lambda_1(t_n)}{\lambda_g(t_n)} = 1 - \frac{\lambda_0(t_n)}{\lambda_g(t_n)}.
\end{equation}
Therefore, the Bayes decision rule is
\begin{equation}
Z^*_n = \argmax_{z \in \{0,1\}} p(Z_n=z|t_n) = \I{\lambda_1(t_n) > \lambda_0(t_n)}
\end{equation}
where $\I{x}=1$ if $x$ is true, and 0 otherwise.
However, this rule cannot be directly applied, because $\lambda_1(t_n)$ is unknown.
From Assumption~\ref{assum:co} (Constant), $\lambda_1$ is a constant, so the decision rule becomes $Z^*_n = \I{\lambda_0(t_n) < \theta_c}$, where $\theta_c$ is a threshold. This justifies ranking by
\begin{equation}
    \Score_c(t_n) = -\lambda_0(t_n)
\end{equation}
across all $n=1,\ldots,N_{\target}$, so we use $-\lambda_0(t_n)$ as the commission outlier score: the higher the score, the more likely $t_n$ is to be a commission outlier.

\paragraph{Nonconstant Rate} We generalize the idea by relaxing the assumption of $\lambda_1(t)$ being a constant. The key is to treat $\lambda_1(t_n)$ as a random variable and then marginalize it out as
\begin{equation}
    \begin{aligned}
        p(Z_n=1|t_n)
        &= \E[\lambda_1]{p(Z_n=1|t_n,\lambda_1(t_n))} \\
        &= 1 + \E[\lambda_1]{\frac{\Score_c(t_n)}{\lambda_1(t_n) - \Score_c(t_n)}}.
    \end{aligned}
\end{equation}
Then we can prove that $p(Z_n=1|t_n)$ is still an increasing function of $\Score_c(t_n)$, which justifies ranking by $\Score_c$ (see Appendix~\ref{sec:nonconstant:commiss} for details).

\subsection{Detecting Omission Outliers}\label{sec:omiss}

Similar to detecting commission outliers, we make the following assumption for omission outliers.
\begin{assumption}[Independent and Constant Omission]\label{assum:oo}
    Omission outliers are generated by independently removing the normal points with a constant probability.  
\end{assumption}

To derive the method, we first define some notations. Let $\lambda_0(t)$ denote the CIF of the normal point process, $p_1$ the probability of removing each normal point, and $B$ a blank interval without any target points which we wish to decide whether contains omission outliers. For any interval $\tau \subseteq \Time$, let $N(\tau)$ be the number of points observed, and $N_0(\tau)$ be the number of points generated by the normal point process with CIF $\lambda_0(t)$, so $N(\cdot)$ is the result of combining $N_0(\cdot)$ with random removal, and we can observe $N(\cdot)$ but not $N_0(\cdot)$. Furthermore, we define an auxiliary random variable $K_B$ that counts the number of points removed in $B$.

For a blank interval $B$, we observe $N(B)=0$, but $K_B = k$ can take different values $k=0, 1, \ldots$. The joint probability for each $k$ is 
\begin{equation} 
    \begin{aligned}
        p(K_B=k, N(B)=0) &= p(K_B=k, N_0(B)=k) \\
        &= p_1^k F_k(B)
    \end{aligned}
\end{equation}
where $F_k(B)$ denotes the probability that $k$ points are generated by the normal point process $N_0(\cdot)$ in $B$ for $k=0,1,\ldots$. It depends on $\lambda_0(t)$.
Then we can calculate the posterior probability of $K_B=0$
\begin{equation}\label{eq:omiss:post}
    \begin{aligned}
        p(K_B=0|N(B)=0)
        =& \frac{F_0(B)}{\sum_{k=0}^\infty p_1^k F_k(B)} 
    \end{aligned}
\end{equation}

Define a random variable $Z_B$ to indicate whether there are any omission outliers in the blank interval $B$. $Z_B = 0$ is equivalent to $K_B = 0$; $Z_B = 1$ is equivalent to $K_B > 0$.
\begin{equation}
    \begin{aligned}
p(Z_B=1|N(B)=0) &= 1 - p(Z_B=0|N(B)=0) \\
&= 1 - p(K_B=0|N(B)=0).
    \end{aligned}
\end{equation}
Then the Bayes decision rule is
\begin{equation}
    \begin{aligned}
Z_B^* &= \argmax_{z\in\{0,1\}} p(Z_B=z|N(B)=0) \\
&= \I{p(K_B=0|N(B)=0) < 0.5}.
    \end{aligned}
\end{equation}

Without further assumptions, $p(K_B=0|N(B)=0)$ (Eq.~\ref{eq:omiss:post}) cannot be evaluated in closed form, but we can get a lower bound
\begin{equation}
    \begin{aligned}
p(K_B=0|N(B)=0) &\ge F_0(B) \\
&= \Exp{-\int_B \lambda_0(s) ds},
    \end{aligned}
\end{equation}
because
\begin{equation*}
    \sum_{k=0}^\infty p_1^k F_k(B) \le \sum_{k=0}^\infty F_k(B) = 1.
\end{equation*}
Then the posterior probability of $B$ containing any omission outliers can also be bounded
\begin{equation}
    \begin{aligned}
        p(Z_B=1|N(B)=0)
        \le & 1 - \Exp{-\int_B\lambda_0(s)ds}.
    \end{aligned}
\end{equation}
Therefore, we propose to use
\begin{equation}
    \Score_o(B) = \int_B\lambda_0(s)ds
\end{equation}
as the omission outlier score. When we rank the blank intervals by $s_o(B)$, we essentially rank them by an upper bound of $p(Z_B=1|N(B)=0)$.

\paragraph{Poisson Process}

There is a notable special case where we can get a closed-form $p(K_B=0|N(B)=0)$.
When the normal point process $N_0(\cdot)$ is an \emph{inhomogeneous Poisson process}, we have
\begin{align}
F_k(B) =& \frac{\left(\int_B \lambda_0(s) ds \right)^k }{k!} \Exp{-\int_B\lambda_0(s)ds}
\end{align}
for $k=0,1,\ldots$. The posterior becomes
\begin{equation}
    \begin{aligned}
p(K_B=0|N(B)=0)
&= \frac{F_0(B)}{\sum_{k=0}^\infty p_1^k F_k(B)} \\
&= \Exp{-p_1 \int_B \lambda_0(s)ds}.
    \end{aligned}
\end{equation}
Therefore, the posterior probability of $B$ containing any omission outliers is
\begin{equation}\label{eq:omiss:post:pois}
    \begin{aligned}
        p(Z_B=1|N(B)=0) = 1 - \Exp{-p_1 \int_B\lambda_0(s)ds}.
    \end{aligned}
\end{equation}
This justifies scoring the interval $B$ by $\Score_o(B)=\int_B\lambda_0(s)ds$, because if we rank the intervals by their scores, the result will be the same as ranking by their posterior probabilities of containing omission outliers, $p(Z_B=1|N(B)=0)$.

\paragraph{Nonconstant Rate}

Similar to commission outliers, we can generalize the idea by relaxing the assumption that $p_1$ is a constant and treating it as a random variable, in which case $p(Z_B=1|N(B)=0)$ is still an increasing function of $\Score_o(B)$ (see Appendix~\ref{sec:nonconstant:omiss} for details).

\paragraph{Inter-Event Time}

Without making any assumptions on the normal point process, we can alternatively justify $\Score_o(B) = \int_B\lambda_0(s)ds$ with hypothesis testing, if $B$ is an \emph{inter-event time} interval, i.e., a time interval between consecutive events $t_{n-1}$ and $t_n$. Let $T_n$ be the random variable for the inter-event time corresponding to $B$. The null and alternative hypotheses are
\begin{equation*}
    H_0: B \text{ is normal}; \quad H_1: B \text{ contains omission outliers}.
\end{equation*}
Assuming the null hypothesis is true, i.e., $B$ is generated by the normal point process with CIF $\lambda_0(t)$, the probability that the inter-event time is at least as long as $|B|$ is
\begin{equation}\label{eq:omiss:pval}
    \begin{aligned}
    p(T_n > |B|) = \Exp{-\int_B\lambda_0(s)ds},
    \end{aligned}
\end{equation}
which is the p-value. A lower p-value means that the observation is more extreme, given that the null hypothesis is true, which means it is more likely to contain omission outliers. This justifies scoring by $\Score_o(B) = \int_B\lambda_0(s)ds$, where a higher score means that $B$ is more likely to contain omission outliers.

\subsection{Bounds on FDR and FPR}\label{sec:bounds}

In this section, we prove some bounds on the performance of the proposed outlier scoring methods. We recall the definitions of false discovery rate (FDR) and false positive rate (FPR). Let $y$ denote the true label (1=outlier, 0=normal) of an object (a target event or a blank interval) and $\hat y$ denote the predicted label.
Then FDR and FPR are defined as 
\begin{equation*}
    FDR = p(y=0|\hat y=1), \quad FPR = p(\hat y=1|y=0).
\end{equation*}
Given the above definitions, we can prove the following theorems (see Appendix~\ref{sec:proof} for proofs).

\begin{theorem}\label{th:commiss:fdr}
    If we use the commission outlier score $\Score_c(t_n) = -\lambda_0(t_n)$, where $t_n$ is the time of a target event, with a threshold $\theta_c \le 0$, such that the decision rule is $\hat y_c(t_n) = \I{\Score_c(t_n) > \theta_c}$, and let $\lambda_1$ denote the CIF of the independent process generating commission outliers, then the FDR is bounded above by $\frac{-\theta_c}{\lambda_1 - \theta_c}$.
\end{theorem}

\begin{theorem}\label{th:omiss:fpr}
    If we use the omission outlier score $\Score_o(B) = \int_B\lambda_0(s)ds$ for an \emph{inter-event time} interval $B$, with a threshold $\theta_o \ge 0$, such that the decision rule is $\hat y_o(B) = \I{\Score_o(B) > \theta_o}$, then the FPR is bounded above by $\Exp{-\theta_o}$.
\end{theorem}

\begin{theorem}\label{th:omiss:fdr}
    If we use the omission outlier score $\Score_o(B) = \int_B\lambda_0(s)ds$ for a blank interval $B$, with a threshold $\theta_o \ge 0$, such that the decision rule is $\hat y_o(B) = \I{\Score_o(B) > \theta_o}$, and assume that the normal point process is an \emph{inhomogeneous Poisson process} and the probability of omission is $p_1$, then the FDR is bounded above by $\Exp{-p_1\theta_o}$.
\end{theorem}

\section{Experiments}

To test the proposed methods, we perform experiments on both synthetic and real-world event sequences. We compare the following methods in the experiments. \textbf{\Rand}: A baseline that generates outlier scores by sampling from a uniform distribution. \textbf{\Len}: A baseline that detects outliers based on the empirical distribution of the inter-event time lengths (see Section~\ref{sec:baseline}). \textbf{\PPOD} (Point-Process Outlier Detection): Our method based on a point-process model but only using the history of the target events as the context. \textbf{\CPPOD} (Contextual PPOD): Our method based on a point-process model using the history of both the target events and the context events as the context.

A model adapted from the continuous-time LSTM \citep{mei_neural_2017} is used for {\PPOD} and {\CPPOD} (see Appendix~\ref{sec:ctlstm}). We choose the number of hidden units in the model from $\{64, 128, 256, 512, 1024\}$ by maximizing the likelihood on the internal validation set that consists of 20 percent of the training set. We stress that for training and validation we do not use any labeled outlier data. Our implementation of all the methods and the experiments is publicly available.\footnote{\url{https://github.com/siqil/CPPOD}}

\subsection{Baseline Method}\label{sec:baseline}

We briefly describe the baseline method {\Len}. For training, the lengths of all the inter-event time intervals of the target events, $L = \{l_i: l_i = t_{i+1} - t_{i}, (t_i,\target), (t_{i+1},\target) \in \St\}$ are collected. Then, an empirical distribution of the inter-event time can be formulated as $
\hat{F}(l) = \frac{1}{|L|} \sum_{i=1}^{|L|} \I{l_i \le l}
$. Here, for simplicity, we describe the method as if we only had one sequence in the training data, but it is easy to see how it works for multiple sequences, which is the case in our experiments.

For testing, {\Len} outputs a commission outlier score for a target event at time $t_n \in \St$ as
\begin{equation}
    \Score_c(t_n) = -\min \{\hat{F}(t_n - t_{n-1}), 1 - \hat{F}(t_n - t_{n-1})\},
\end{equation}
where $t_n - t_{n-1}$ is the inter-event time between the current and previous target events. Intuitively, if the inter-event time is too small ($\hat{F}(\cdot)$ is small) or too big ($1-\hat{F}(\cdot)$ is small), it is likely that $t_n$ has occurred at an abnormal time (too early or too late) and therefore is a commission outlier. The negation makes sure that a higher score indicates that it is more likely to be an outlier. For a blank interval $B$, {\Len} outputs an omission outlier score as the length of $B$,
\begin{equation}
    \Score_o(B) = |B|.
\end{equation}
Intuitively, the longer the blank interval, the more likely it is to contain omission outliers.

\subsection{Experiments on Synthetic Event Sequences}\label{sec:exp:synth}

We generate synthetic event sequences using two different types of point processes. One is the inhomogeneous Poisson process. The other is the Gamma process. For each type of processes, there is a set of parameters that determine the distribution of the points. We allow the parameters to vary according to a context state $x$, which can take two distinct values $x \in \{0, 1\}$. 

For the inhomogeneous Poisson process, the CIF is a piecewise constant function with the value $\lambda = f(x)$, where $x$ is the context state. In the experiments, we set $f(0) = 0.1$ and $f(1) = 1$.
For the Gamma process, the inter-event time follows a Gamma distribution $\Gam{a_x, b_x}$ ($a_x$ shape, $b_x$ rate), where $x$ is the context state. In the experiments, we set $(a_0,b_0) = (10, 10)$ and $(a_1,b_1) = (100, 10)$.

The changes of the context state $x$ are driven by a continuous-time Markov chain \citep{nodelman_continuous_2002} with a transition matrix
$
    Q =
    \begin{bmatrix}
        -0.05 & 0.05 \\
        0.05 & -0.05 \\
    \end{bmatrix}
$
such that
$$
    p\left(x(t+dt) = j|x(t) = i\right) = \I{i=j} + Q_{ij} dt,
$$
where $dt$ is infinitesimal time. Each change of the state generates a context event.

For each point process type, we simulate 40 sequences. All sequences are simulated in the same time span $\Time = [0, 1000]$. We use 50 percent of the sequences for training and the others for testing.

\paragraph{Outlier Simulation} Commission outliers are simulated by adding points generated from a separate point process. Omission outliers are simulated by random removals of existing points. There is a parameter $\alpha(t)$ that defines the rate of outliers relative to normal points and can change over time (see Appendix~\ref{sec:sim} for details). To evaluate the performance of the proposed methods under different conditions, we conduct three types of simulations.
\begin{itemize}
    \item \textbf{Constant rate.} In this case, the rate of omission and commission is a constant, $\alpha(t) = \alpha_0$, relative to the normal points, where $\alpha_0 = 0.1$. We also changed $\alpha_0$, but it did not affect the relative performance of each method in almost all cases (see Appendix~\ref{sec:additional:results}).
    \item \textbf{Periodic rate} (denoted as [sin]). In this case, the rate of omission and commission is a periodic function defined by $\alpha(t) = \alpha_0 (1+\sin(2\pi t / p))/2$ relative to the normal points, where $\alpha_0$ controls the overall scale, $t$ is the time, and $p$ is the period. We set $\alpha_0 = 0.2$ and $p = 100$.
    \item \textbf{Piecewise-constant rate} (denoted as [pc]). In this case, we randomly generate a piecewise-constant function $g(t): \Time \to [0, 1]$ with a step size $s = 10$ for each sequence. The value of the function at each step is randomly generated from a uniform distribution. Then the rate of omission and commission is defined as $\alpha(t) = \alpha_0 g(t)$ relative to the normal points, where $\alpha_0 = 0.2$ controls the overall scale.
\end{itemize}
We note that the first type satisfies the constant rate assumptions we made when developing the methods, but the second and third types do not. By using different simulations, we try to verify the robustness of our methods in different cases when the constant rate assumptions may not be satisfied.

\paragraph{Outlier Detection} We apply the methods to detect outliers in an online manner, i.e., a commission (omission) outlier score is generated whenever a new target event is observed. For omission outliers, we also try to detect outliers at additional checkpoints within long blank intervals. See Appendix~\ref{sec:detect} for details of the approach and Appendix~\ref{sec:outlier:ratio} for the outlier ratios of the datasets.

\paragraph{Results}

\begin{table*}[!htbp]
    \centering
    \caption{AUROC on synthetic data. Dataset: name abbreviation (C=commission, O=omission) [$\alpha$].}\label{tab:roc:synthetic:main}
    \begin{tabular}{l|lll|lll}
        \toprule
        Dataset & Poi (C) [0.1]              & Poi (C) [sin]              & Poi (C) [pc]               & Poi (O) [0.1]              & Poi (O) [sin]              & Poi (O) [pc]               \\
\hline
\Rand   & .500 ($\pm$ .010)          & .493 ($\pm$ .007)          & .512 ($\pm$ .009)          & .503 ($\pm$ .008)          & .498 ($\pm$ .013)          & .491 ($\pm$ .007)          \\
\Len    & .601 ($\pm$ .008)          & .575 ($\pm$ .006)          & .584 ($\pm$ .011)          & .650 ($\pm$ .006)          & .659 ($\pm$ .007)          & .652 ($\pm$ .011)          \\
\PPOD   & .684 ($\pm$ .010)          & .661 ($\pm$ .016)          & .664 ($\pm$ .009)          & .737 ($\pm$ .006)          & .741 ($\pm$ .012)          & .734 ($\pm$ .013)          \\
\CPPOD  & \textbf{.711} ($\pm$ .012) & \textbf{.707} ($\pm$ .017) & \textbf{.697} ($\pm$ .014) & \textbf{.778} ($\pm$ .005) & \textbf{.791} ($\pm$ .010) & \textbf{.784} ($\pm$ .010) \\
\hline
Dataset & Gam (C) [0.1]              & Gam (C) [sin]              & Gam (C) [pc]               & Gam (O) [0.1]              & Gam (O) [sin]              & Gam (O) [pc]               \\
\hline
\Rand   & .485 ($\pm$ .007)          & .493 ($\pm$ .008)          & .506 ($\pm$ .007)          & .505 ($\pm$ .012)          & .503 ($\pm$ .010)          & .515 ($\pm$ .010)          \\
\Len    & .754 ($\pm$ .006)          & .762 ($\pm$ .008)          & .757 ($\pm$ .005)          & .799 ($\pm$ .005)          & .809 ($\pm$ .006)          & .813 ($\pm$ .005)          \\
\PPOD   & .816 ($\pm$ .008)          & .817 ($\pm$ .006)          & .813 ($\pm$ .005)          & .901 ($\pm$ .007)          & .902 ($\pm$ .006)          & .905 ($\pm$ .006)          \\
\CPPOD  & \textbf{.871} ($\pm$ .006) & \textbf{.886} ($\pm$ .004) & \textbf{.870} ($\pm$ .007) & \textbf{.956} ($\pm$ .003) & \textbf{.956} ($\pm$ .004) & \textbf{.955} ($\pm$ .004) \\

        \bottomrule
    \end{tabular}
\end{table*}

Table~\ref{tab:roc:synthetic:main} shows the results in Area Under the Receiver Operating Characteristic curve (AUROC). {\CPPOD} achieves the best performance for both commission and omission outliers, showing the effectiveness of our outlier scoring methods. {\PPOD} being worse than {\CPPOD} shows the importance of the context events in these cases. {\Len} performs much better than {\Rand} but worse than our proposed methods, which shows (1) it is an effective method for detecting outliers; (2) although intuitive, it is less effective than our proposed methods. Meanwhile, it does not have rigorous theoretical justifications.

As we can see, changing the outlier simulation mechanism does not affect the relative performance of each method. Even when the rates are not constants, i.e., $\alpha = \text{[sin]}$ or $\text{[pc]}$, {\CPPOD} and {\PPOD} are still performing similarly as when $\alpha$ is a constant and much better than the baselines. We also checked the performance of our outlier scoring methods combined with the ground-truth model (see Appendix~\ref{sec:additional:results}), and the performance is similar to {\CPPOD}.  

\paragraph{Empirical Verification of the Bounds on FDR and FPR}

To empirically verify the bounds on FDR and FPR presented in Section~\ref{sec:bounds}, we randomly repeat the experiments using our outlier scoring methods with the ground-truth model  (replacing the learned LSTM model with the true model in {\CPPOD}) on the synthetic data 10 times with different test data. Each time we calculate the FDR and FPR for different thresholds on the scores. For verifying FPR, we only test the inter-event time intervals for omission outliers. Their means and standard deviations over all repetitions are shown with the theoretical bounds in Figure~\ref{fig:bound:gam} for Gamma process. For FPR, the bounds overlap the empirical rates. See Appendix~\ref{sec:bound:pois} for more details and the results for Poisson process.

\begin{figure}[!htbp]
    \centering
    \includegraphics[width=0.49\linewidth]{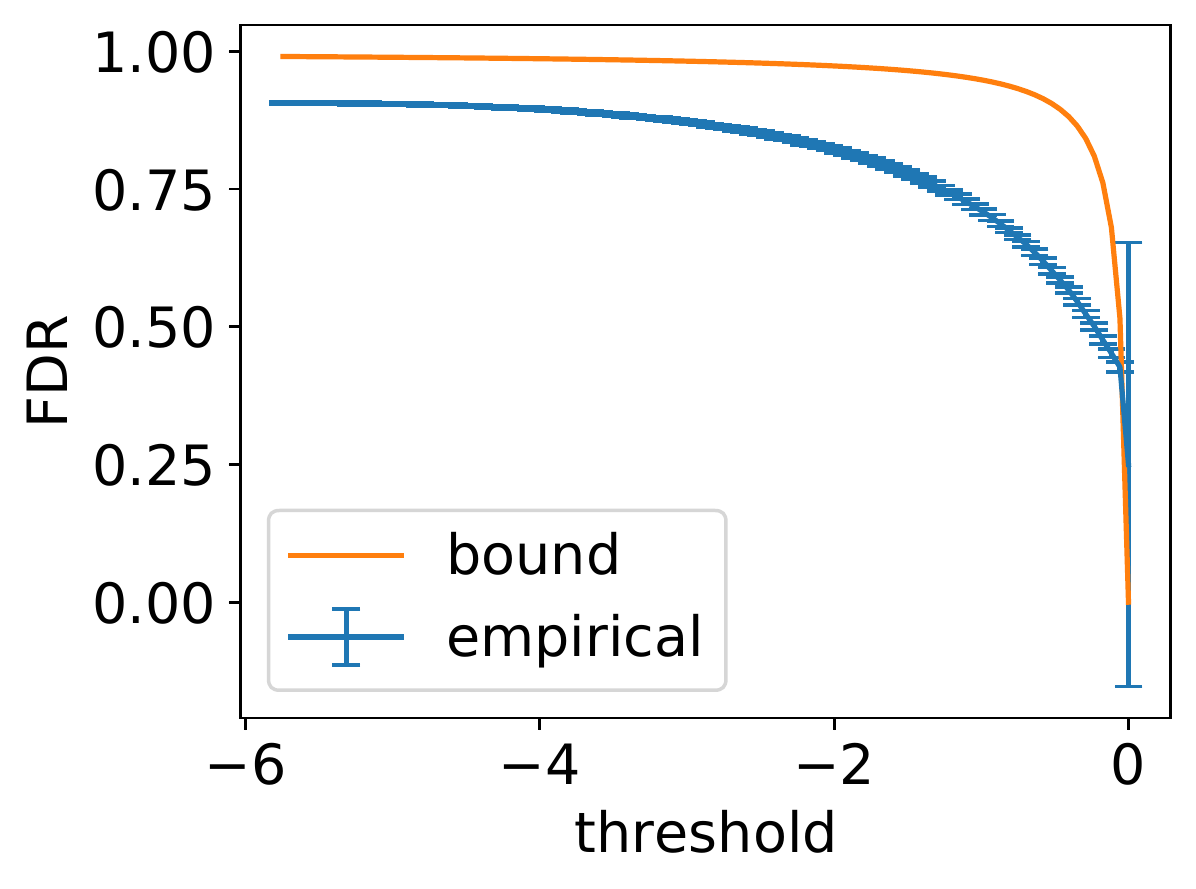}
    \includegraphics[width=0.49\linewidth]{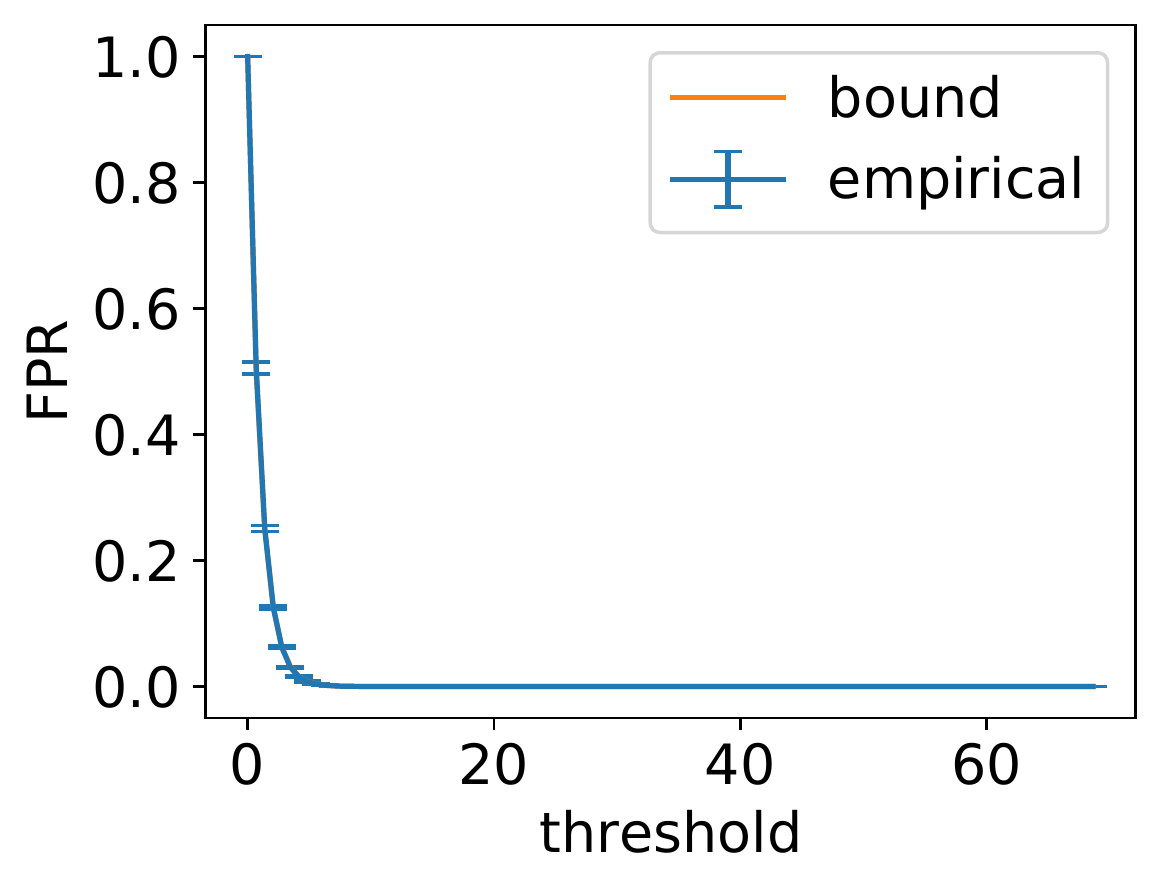}
    \caption{FDR (commission outlier) and FPR (omission outlier) on synthetic data (Gamma process).}\label{fig:bound:gam}
\end{figure}

\subsection{Experiments on Real-World Clinical Data}

In this part, we use real-world clinical data extracted from the MIMIC III dataset \citep{johnson_mimic-iii_2016}. The dataset consists of de-identified electronic health records of ICU patients. We use four types of events corresponding to commonly used medications and lab tests as our targets and form four datasets by collecting the target events and corresponding context events. The target events and their context events are listed in Table~\ref{tab:mimic:name}. The medical category (medication, lab, or vital sign) of each type of events is in brackets following the type. For example, Potassium Chloride is a type of medication, and Potassium (Blood) is a type of lab test. The latter is used as the context for the former, as the administration of the medication can be triggered by observing an abnormally low value in the lab test.

We record in the data every type of events in the table. However, for Potassium (Blood) and Total Calcium (Blood), we further split the context events into three subtypes depending on whether the value in the lab test is low, normal, or high. For INR(PT) (previous state), we create context events of two subtypes based on whether the value of the \emph{previous} event is normal or abnormal (with a 1-second delay). For Arterial Blood Pressure systolic (ABPs) and Non-invasive Blood Pressure systolic (NBPs), we split the events into two subtypes depending on whether the value is normal or low. These subtypes help to define better the contexts influencing the target events, since depending on their values, the target events can be more/less likely to occur. 

All target and context events for one patient admission form one event sequence. We randomly select 2000 sequences for the first three datasets and 500 sequences for the last one, as each sequence in the last one contains much more events than the first three. For each dataset, we use 50 percent of the sequences for training and the others for testing.

\begin{table}[!htbp]
    \centering
    \caption{Names of target and context events from MIMIC. INR=international normalized ratio; PT=prothrombin time.}\label{tab:mimic:name}
    \begin{tabular}{p{2.8cm}|p{4.8cm}}
        \toprule
        Target & Context\\
        \midrule
        INR(PT) [Lab] & INR(PT) [Lab] (previous state); \newline Heparin [Medication]; \newline Warfarin [Medication] \\
        \hline
        Calcium Gluconate [Medication] & Total Calcium (Blood) [Lab] \\
        \hline
        Potassium Chloride [Medication] & Potassium (Blood) [Lab] \\
        \hline
        Norepinephrine [Medication] & Arterial Blood Pressure systolic [Vital Sign]; \newline Non-invasive Blood Pressure systolic [Vital Sign] \\
        \bottomrule
    \end{tabular}
\end{table}

\paragraph{Simulation Evaluation}

\begin{table*}[!htbp]
    \centering
    \caption{AUROC on MIMIC data. Dataset: name abbreviation (C=commission, O=omission) [$\alpha$].}\label{tab:roc:mimic:main}
    \begin{tabular}{l|lll|lll}
        \toprule
        Dataset & INR (C) [0.1]                & INR (C) [sin]                & INR (C) [pc]                 & INR (O) [0.1]                & INR (O) [sin]                & INR (O) [pc] \\
\hline
\Rand   & .496 ($\pm$ .010)          & .508 ($\pm$ .009)          & .488 ($\pm$ .010)          & .498 ($\pm$ .011)          & .516 ($\pm$ .012)          & .508 ($\pm$ .009) \\
\Len    & .596 ($\pm$ .009)          & .588 ($\pm$ .010)          & .607 ($\pm$ .010)          & .726 ($\pm$ .008)          & .717 ($\pm$ .011)          & .720 ($\pm$ .011) \\
\PPOD   & .682 ($\pm$ .010)          & .675 ($\pm$ .009)          & .673 ($\pm$ .008)          & \textbf{.748} ($\pm$ .009) & .760 ($\pm$ .010)          & \textbf{.773} ($\pm$ .009) \\
\CPPOD  & \textbf{.687} ($\pm$ .009) & \textbf{.680} ($\pm$ .009) & \textbf{.681} ($\pm$ .010) & .746 ($\pm$ .010)          & \textbf{.764} ($\pm$ .009) & .770 ($\pm$ .009) \\
\hline
Dataset & Cal (C) [0.1]                & Cal (C) [sin]                & Cal (C) [pc]                 & Cal (O) [0.1]                & Cal (O) [sin]                & Cal (O) [pc] \\
\hline
\Rand   & .504 ($\pm$ .013)          & .502 ($\pm$ .016)          & .508 ($\pm$ .011)          & .493 ($\pm$ .016)          & .518 ($\pm$ .017)          & .496 ($\pm$ .017) \\
\Len    & .739 ($\pm$ .012)          & .688 ($\pm$ .015)          & .742 ($\pm$ .011)          & .526 ($\pm$ .009)          & .529 ($\pm$ .012)          & .541 ($\pm$ .010) \\
\PPOD   & .830 ($\pm$ .010)          & .797 ($\pm$ .010)          & .837 ($\pm$ .009)          & .759 ($\pm$ .008)          & .758 ($\pm$ .009)          & .759 ($\pm$ .011) \\
\CPPOD  & \textbf{.866} ($\pm$ .006) & \textbf{.835} ($\pm$ .009) & \textbf{.860} ($\pm$ .011) & \textbf{.775} ($\pm$ .008) & \textbf{.777} ($\pm$ .010) & \textbf{.780} ($\pm$ .009) \\
\hline
Dataset & Pot (C) [0.1]                & Pot (C) [sin]                & Pot (C) [pc]                 & Pot (O) [0.1]                & Pot (O) [sin]                & Pot (O) [pc] \\
\hline
\Rand   & .498 ($\pm$ .012)          & .503 ($\pm$ .010)          & .511 ($\pm$ .010)          & .495 ($\pm$ .017)          & .508 ($\pm$ .011)          & .524 ($\pm$ .010) \\
\Len    & .733 ($\pm$ .013)          & .691 ($\pm$ .009)          & .718 ($\pm$ .013)          & .533 ($\pm$ .012)          & .536 ($\pm$ .014)          & .552 ($\pm$ .011) \\
\PPOD   & .839 ($\pm$ .009)          & .813 ($\pm$ .011)          & .831 ($\pm$ .008)          & .736 ($\pm$ .011)          & .744 ($\pm$ .011)          & .746 ($\pm$ .011) \\
\CPPOD  & \textbf{.878} ($\pm$ .009) & \textbf{.857} ($\pm$ .007) & \textbf{.874} ($\pm$ .010) & \textbf{.748} ($\pm$ .011) & \textbf{.759} ($\pm$ .012) & \textbf{.761} ($\pm$ .011) \\
\hline
Dataset & Nor (C) [0.1]                & Nor (C) [sin]                & Nor (C) [pc]                 & Nor (O) [0.1]                & Nor (O) [sin]                & Nor (O) [pc] \\
\hline
\Rand   & .494 ($\pm$ .014)          & .536 ($\pm$ .012)          & .524 ($\pm$ .012)          & .510 ($\pm$ .010)          & .488 ($\pm$ .014)          & .503 ($\pm$ .012) \\
\Len    & .864 ($\pm$ .010)          & .837 ($\pm$ .012)          & .844 ($\pm$ .016)          & .468 ($\pm$ .016)          & .462 ($\pm$ .013)          & .476 ($\pm$ .014) \\
\PPOD   & .890 ($\pm$ .012)          & .858 ($\pm$ .014)          & \textbf{.884} ($\pm$ .014) & \textbf{.835} ($\pm$ .010) & \textbf{.842} ($\pm$ .011) & \textbf{.851} ($\pm$ .011) \\
\CPPOD  & \textbf{.897} ($\pm$ .013) & \textbf{.871} ($\pm$ .014) & .882 ($\pm$ .013)          & .832 ($\pm$ .009)          & .837 ($\pm$ .011)          & .848 ($\pm$ .010) \\

        \bottomrule
    \end{tabular}
\end{table*}

We generate and detect commission and omission outliers on top of the existing data using the same procedures for synthetic data (Section~\ref{sec:exp:synth}) except that we set $p = 24 \times 7$ and $s = 12$.
This allows us to obtain ground-truth labels for analyses.
Table~\ref{tab:roc:mimic:main} shows the AUROC results.
The results have more variations across different datasets in this case, which can be seen by examining the performance of {\Len}.
Omission outliers appear to be more challenging than commission outliers except for INR(PT).
{\CPPOD} and {\PPOD} outperform {\Rand} and {\Len} on all the datasets for both types of outliers.

In all cases, {\CPPOD} is the best or very close to it. In the latter cases, the best is always {\PPOD}, and the differences are very small. These are the cases where the additional context events are not as influential as the history of the target events themselves for the occurrences of the target events, so {\PPOD} is as good as but simpler than {\CPPOD}. However, for Potassium Chloride and Calcium Gluconate, we can see a clear advantage of {\CPPOD} over {\PPOD} by using additional context events.

\paragraph{Manual Evaluation of Clinical Relevance}

\begin{table}[!htbp]
    \centering
    \caption{Manual evaluation on MIMIC data.}\label{tab:tally:mimic:manual}
    \begin{tabular}{c|c|c|c|c|c|c|c|c}
        \toprule
        Target & \multicolumn{2}{c|}{INR} & \multicolumn{2}{c|}{Cal} & \multicolumn{2}{c|}{Pot} & \multicolumn{2}{c}{Nor} \\
        \hline
        Outlier & C & O & C & O & C & O & C & O \\
        \midrule
        Y & 1 & 2 & 1 & 3 & 1 & 2 & 1 & 2 \\
        P & 1 & 1 & 0 & 0 & 0 & 0 & 1 & 0 \\
        N & 1 & 0 & 2 & 0 & 2 & 1 & 1 & 1 \\
        \bottomrule
    \end{tabular}
\end{table}

Finally, we apply {\CPPOD} to the original data (without any simulated outliers). For each dataset, we select three commission outliers and three omission outliers with the highest outlier scores. We manually examine the electronic health record of the patient for each outlier to determine whether the event commissions and omissions identified by {\CPPOD} are clinically relevant and assign the outliers to three classes: N (outlier is not relevant), Y (outlier is relevant), and P (outlier is probably relevant given the available data but cannot be determined due to insufficient data recorded in MIMIC).

Specifically, we retrieve medical notes and related data for each case across time. We examine the notes and data to find evidence to support the decisions made by the physicians (giving a medication / lab test or not) that are detected as outliers. If there is evidence supporting the decision (e.g., given the condition of the patient, the medication should be given, and it was given), then the outlier is not clinically relevant (N). If the evidence is against the decision, it is clinically relevant (Y). If we do not have enough data, but the data we have do not support the decision, it is probably relevant (P).

Table~\ref{tab:tally:mimic:manual} shows the results of the manual evaluation. 
Overall, {\CPPOD} is performing well, as many detected outliers are clinically relevant. For both false outliers in Cal (C), there was an apparent change in the schedule of the medication (previous pattern being interrupted), so for the model, they do appear to be outliers. For Pot (C), one patient had rare conditions, for which we do not have the corresponding context events. The other had a drop in the value of the lab test (context), but it did not reach the \emph{abnormally} low level, so the model does not have the exact context information that supported the human decision of giving the medication.

We note that for this evaluation, we do not have any guarantees on any of the assumptions we made when developing the methods in theory. Nonetheless, {\CPPOD} is still effective. Moreover, as we have discovered, the original data may already contain outliers, so even in the previous evaluation, the models are unlikely to be trained on clean data, but Table~\ref{tab:roc:mimic:main} still shows the benefits of {\PPOD} and {\CPPOD}. This demonstrates that the assumptions do not have to be rigorously satisfied in practice for the methods to work reasonably well, although we may lose the theoretical guarantees on the performance, which depend on the assumptions.

\section{Discussion}

In this work, we have studied two new outlier detection problems: detection of commission and omission outliers in continuous-time event sequences. We proposed outlier scoring methods based on Bayesian decision theory and hypothesis testing with theoretical guarantees. The proposed methods depend on a point-process model built from the data. In this work, we experimented with a model considering context, adapted from the continuous-time LSTM. We conducted experiments on both synthetic and real-world event sequences. The results show the flexibility of the adapted model and, more importantly, the effectiveness of the proposed outlier scoring methods.

As verified by the experiments, the assumptions we made when developing the methods in theory do not have to be satisfied for them to work reasonably well in practice, although when they are satisfied, we get the theoretical justifications and guarantees on the performance. For future work, there are two interesting directions. One is to develop algorithms to learn the models robustly in presence of outliers in the training data. The other is to allow the model to adapt in a non-stationary environment. Both are orthogonal to this work, as they can be combined with the outlier scoring methods developed in this work to improve the performance by providing a better model.

\section*{Acknowledgement}

This work was supported by NIH grant R01-GM088224. The content of this paper is solely the responsibility of the authors and does not necessarily represent the official views of the NIH. Siqi Liu also acknowledges the support by CS50 Merit Pre-doctoral Fellowship from the Department of Computer Science, University of Pittsburgh. Finally, the authors would like to thank all the anonymous reviewers who provided helpful comments on the earlier versions of the paper.

\bibliography{pp,outlier}
\bibliographystyle{icml2021}

\clearpage

\appendix

\numberwithin{equation}{section}
\numberwithin{figure}{section}
\numberwithin{table}{section}

\section{Continuous-Time LSTM with Context}\label{sec:ctlstm}

The input to the continuous-time LSTM consists of the marked events in the combined sequence, $(t_i, u_i) \in \Sm$. That is, we not only use the target events but also the context events as input, although we only model the CIF of the target events, $\lambda(t)$. The output consists of the hidden states $\V h(t_i)$ corresponding to the input. It is a nonlinear mapping from the content in the memory cell $\V c(t_i)$ of the LSTM at time $t_i$. As in a traditional LSTM, each continuous-time LSTM unit also has an input gate $\V i$, an output gate $\V o$, and a forget gate $\V f$. The relations between the memory cells, the hidden states, the input, and these gates are summarized as follows.

Let $\V u_i$ be a vector representation of the mark $u_i$, which is a learnable embedding. For $t \in (t_{i-1}, t_i]$, $\V c(t)$ is a continuous function changing over time from $\V c_i$ to $\bar{\V c}_i$, and for $\V c_i$ and $\bar{\V c}_i$ there are separate input gates and forget gates:
\begin{align}
    &\V h(t) = \V o_i \odot \tanh(\V c(t)) \\
    &\V c(t) = \bar{\V c}_{i} + (\V c_{i} - \bar{\V c}_{i})\Exp{-\V \delta_{i}(t-t_{i-1})} \\
    &[\V i_{i+1}; \V o_{i+1}; \V f_{i+1}] = \sigma(\V W \V u_i + \V U \V h(t_i) + \V d) \\
    &[\bar{\V i}_{i+i}; \bar{\V f}_{i+1}] = \sigma(\bar{\V W} \V u_i + \bar{\V U} \V h(t_i) + \bar{\V d}) \\
    &\V z_{i+1} = \tanh(\V W_z \V u_i + \V U_z \V h(t_i) + \V d_z) \\
    &\V c_{i+1} = \V f_{i+1} \odot \V c(t_i) + \V i_{i+1} \odot \V z_{i+1} \\
    &\bar{\V c}_{i+1} = \bar{\V f}_{i+1} \odot \bar{\V c_i} + \bar{\V i}_{i+1} \odot \V z_{i+1} \\
    &\V \delta_{i+1} = g(\V W_{\delta} \V u_i + \V U_{\delta} \V h(t_i) + \V d_{\delta}, 1)
\end{align}
where $[\V a;\V b]$ denotes the concatenation of the vectors $\V a$ and $\V b$, $\odot$ is the elementwise product, $\sigma(\cdot)$ is the logistic function, and $g(x, s) = s\log(1+\Exp{x/s})$ is the scaled softplus function with parameter $s$. All the $\V W$, $\V U$ and $\V d$ with/without different subscripts and bars are learnable parameters of the continuous-time LSTM.

Finally, to convert the output of the continuous-time LSTM to the CIF of the target events, $\lambda(t)$, we have
$\lambda(t) = g(\V w_\lambda^T \V h(t), s)$
where $\V w_\lambda$ and $s$ are learnable parameters. The model is learned by maximizing the likelihood (Eq.~\ref{eq:pdf}) for all sequences in the training data. Monte-Carlo integration is used to evaluate $\int\lambda(s)ds$.

\section{Generalization to Nonconstant Commission}\label{sec:nonconstant:commiss}

We generalize the method we have developed in Section~\ref{sec:commiss} to cases when the rate of commission is not a constant. Treating $\lambda_1(t_n)$ as a random variable, based on what we have already developed, we have
\begin{align*}
    p(Z_n=1|t_n,\lambda_1(t_n)) &= 1 - \frac{\lambda_0(t_n)}{\lambda_g(t_n)} \\
    &= 1 - \frac{\lambda_0(t_n)}{\lambda_0(t_n)+\lambda_1(t_n)} \\
    &= 1 + \frac{\Score_c(t_n)}{\lambda_1(t_n) - \Score_c(t_n)} \\
\end{align*}
To avoid cluttering, we omit $t_n$ in $\lambda_0$, $\lambda_1$, and $\Score_c$ from now. By marginalizing out $\lambda_1$, we get
\begin{align*}
        p(Z_n=1|t_n)
        &= \E[\lambda_1]{p(Z_n=1|t_n,\lambda_1)} \\
        &= 1 + \E[\lambda_1]{\frac{\Score_c}{\lambda_1 - \Score_c}} \\
        &= 1 + \E[\lambda_1]{f(\Score_c, \lambda_1)}
\end{align*}
where we defined
\[
    f(\Score_c, \lambda_1) = \frac{\Score_c}{\lambda_1 - \Score_c}
\]

We assume
\[
    \E[\lambda_1]{\frac{1}{\lambda_1}} < \infty
\]
which is easy to satisfy, since it is sufficient that either $\lambda_1 \ge \epsilon$ for some $\epsilon > 0$ or the distribution of $\lambda_1$ is one of the common distributions including any finite discrete distribution, Gamma distribution, etc.

It is not hard to see that $f$ is an increasing function of $\Score_c$ for any given $\lambda_1$, as $\Score_c \in (-\infty, 0)$, $\lambda_1 \in (0, \infty)$, and
\[
    \PD{f}{\Score_c} = \frac{\lambda_1 - 2\Score_c}{(\lambda_1 - \Score_c)^2} > 0
\]
Meanwhile,
\[
    \abs{\PD{f}{\Score_c}} = \abs{\frac{\lambda_1 - 2\Score_c}{(\lambda_1 - \Score_c)^2}} < \frac{1}{\lambda_1}
\]
Therefore, we can show that
\[
    \PD{\E[\lambda_1]{f(\Score_c, \lambda_1)}}{\Score_c} = \E[\lambda_1]{\PD{f(\Score_c, \lambda_1)}{\Score_c}} > 0
\]
using the Dominated Convergence Theorem. This implies that $p(Z_n=1|t_n)$ is an increasing function of $s_c$.

\section{Generalization to Nonconstant Omission}\label{sec:nonconstant:omiss}

Similar to commission outliers, we generalize the method we have developed in Section~\ref{sec:omiss} to cases when the probability of omission is not a constant and the normal point process is an inhomogeneous Poisson process. Treating $p_1$ as a random variable, based on what we have already developed, we have
\begin{align*}
    p(Z_B=1|N(B)=0,p_1)
    &= 1 - \Exp{-p_1 \int_B\lambda_0(s)ds} \\
    &= 1 - \Exp{-p_1 \Score_o(B)}
\end{align*}

To avoid cluttering, we omit $B$ in $\Score_o$ from now. By marginalizing out $p_1$, we get
\begin{align*}
    p(Z_B=1|N(B)=0)
    &= 1 - \E[p_1]{\Exp{-p_1 \Score_o}} \\
    &= 1 - \E[p_1]{g(\Score_o, p_1)}
\end{align*}
where we defined
\[
    g(\Score_o, p_1) = \Exp{-p_1 \Score_o}
\]

Apparently $g$ is an decreasing function of $\Score_c$ for any given $p_1$, as $\Score_o \in (0, \infty)$, $p_1 \in [0, 1]$, and
\[
    \PD{g}{\Score_o} = (-p_1)\Exp{-p_1 \Score_o} \le 0
\]
Meanwhile,
\[
    \abs{\PD{g}{\Score_o}} = \abs{(-p_1)\Exp{-p_1 \Score_o}} \le 1
\]
Therefore, we can show that
\[
    \PD{\E[p_1]{g(\Score_o, p_1)}}{\Score_o} = \E[p_1]{\PD{g(\Score_o, p_1)}{\Score_o}} \le 0
\]
using the Dominated Convergence Theorem. This implies that $p(Z_B=1|N(B)=0)$ is an increasing function of $s_o$.

\section{Proofs of the Theorems}\label{sec:proof}

\subsection{Theorem~\ref{th:commiss:fdr}}

\begin{proof}
    From Eq.~\ref{eq:commiss:post} and implicitly conditioned on the event $t_n$ and the history
    \begin{equation*}
        \begin{aligned}
            p(y_c(t_n)=0) = p(Z_n=0) = \frac{\lambda_0(t_n)}{\lambda_0(t_n)+\lambda_1}
        \end{aligned}
    \end{equation*}
    Given that $\hat y_c(t_n) = 1$, i.e., $-\lambda_0(t_n) > \theta_c$, we get
    \begin{equation*}
        p(y_c(t_n)=0|\hat y_c(t_n)=1) < \frac{-\theta_c}{-\theta_c+\lambda_1}
    \end{equation*}
\end{proof}

\subsection{Theorem~\ref{th:omiss:fpr}}

\begin{proof}
    Let $T_n$ be the random variable for the inter-event time corresponding to the observed inter-event interval $B$, assuming it is generated from the normal point process. From Eq.~\ref{eq:omiss:pval}
    \begin{equation*}
        \begin{aligned}
            & p(\hat y_o(B)=1|y_o(B)=0) \\
            =& p \left(\int_B\lambda_0(s)ds > \theta_o \middle| y_o(B)=0 \right) \\
            =& p \left(\Exp{-\int_B\lambda_0(s)ds} < \Exp{-\theta_o} \middle| y_o(B)=0 \right) \\
            =& p \left(p(T_n > |B|) < \Exp{-\theta_o} \right) \\
            =& \Exp{-\theta_o}
        \end{aligned}
    \end{equation*}
    The last equality is because $p(T_n > |B|) = 1 - p(T_n \le |B|)$, and $p(T_n \le |B|)$ is the cumulative distribution function of $T_n$, implying it follows a uniform distribution.
\end{proof}

\subsection{Theorem~\ref{th:omiss:fdr}}

\begin{proof}
    From Eq.~\ref{eq:omiss:post:pois} and implicitly conditioned on $N(B) = 0$ and the history
    \begin{equation*}
        \begin{aligned}
            p(y_o(B)=0) = p(K_B=0) = \Exp{-p_1\int_B\lambda_0(s)ds}
        \end{aligned}
    \end{equation*}
    Given that $\hat y_o(B) = 1$, i.e., $\int_B\lambda_0(s)ds > \theta_o$, we get
    \begin{equation*}
        \begin{aligned}
            p(y_o(B)=0|\hat y_o(B)=1) < \Exp{-p_1\theta_o}
        \end{aligned}
    \end{equation*}    
\end{proof}

\section{Simulation of Commission and Omission Outliers}\label{sec:sim}

To define outliers, we simulate commission and omission outliers on top of the existing data. In this way, we can obtain ground-truth labels for testing.

To define commission outliers, we simulate a new sequence of target events independently from the existing data, and then merge the new events with the existing events. We use an (inhomogeneous) Poisson process with an intensity $\lambda_c(t)$ to generate the outliers. $\lambda_c$ controls the rate of such outliers. In the experiments, for each dataset, we set $\lambda_c(t) = \alpha(t) \hat{\lambda}_{test}$, where $\alpha(t)$ is either a constant or a function over time depending on the settings, and $\hat{\lambda}_{test}$ is the empirical rate of the target events calculated from the original test data.

To define omission outliers, we randomly remove target events in the original sequences according to independent Bernoulli trials. That is, each event is removed with probability $p_1$ and kept with probability $1-p_1$. We always keep the event if it marks the start time of the sequence. In the experiments, we set $p_1 = \alpha(t)$, where, similar to commission, $\alpha(t)$ is either a constant or a function over time depending on the settings.

\section{Detection of Commission and Omission Outliers}\label{sec:detect}

We detect the presence of commission and omission outliers differently. To test for commission outliers, each method outputs an outlier score at the time of each target event. That is, whenever there is a new target event, we ask the question: is this event a commission outlier or not?

Testing for omission outliers is trickier, because we need to decide the checkpoints more carefully, i.e., when to ask for outlier scores. The simplest thing to do is to only check at the target event times. That is, whenever there is a new target event, we ask the question: is there any omission outlier since the previous target event till now?

However, this may become unsatisfactory in real-world applications, because there could be cases when the target events just stop occurring for a long period of time or even forever (potentially due to malfunctions of the underlying system). These are interesting and important cases we are supposed to detect, but the above testing method will not work. Therefore, we use a combined approach. We still have a checkpoint at each target event time, but on top of that, we also randomly generate checkpoints in long blank intervals.

Specifically, we have a parameter $w$ set to $2/\hat{\lambda}_{train}$, where $\hat{\lambda}_{train}$ is the empirical rate of the target events estimated from the training data for each dataset, so within $w$, on average, we should see two events normally. Then, whenever the blank interval from the previous checkpoint till now is longer than $w$, we generate a new checkpoint within the interval by uniform sampling, and set the previous checkpoint to the generated checkpoint. We keep generating checkpoints until we reach the next target event or the end of the sequence.

\section{Outlier Ratios}\label{sec:outlier:ratio}

The outlier ratio, i.e., the number of outliers divided by the total number of test points, for each dataset is summarized in Table~\ref{tab:outlier:ratio}.

\begin{table}[ht]
    \setlength{\tabcolsep}{4pt}
    \centering
    \caption{Outlier ratios of the datasets. Dataset: name abbreviation (C=commission, O=omission) [$\alpha$].}
    \label{tab:outlier:ratio}
    \begin{tabular}{lll|l|lll|l}
        \toprule
        \multicolumn{3}{c|}{Dataset} &  Ratio & \multicolumn{3}{c|}{Dataset} &  Ratio \\
        \midrule
        Gam & (C) & [0.05] & 0.047 & Gam & (O) & [0.05] & 0.034 \\
Gam & (C) & [0.1]  & 0.095 & Gam & (O) & [0.1]  & 0.072 \\
Gam & (C) & [sin]  & 0.089 & Gam & (O) & [sin]  & 0.069 \\
Gam & (C) & [pc]   & 0.088 & Gam & (O) & [pc]   & 0.067 \\
\hline
Poi & (C) & [0.05] & 0.046 & Poi & (O) & [0.05] & 0.033 \\
Poi & (C) & [0.1]  & 0.092 & Poi & (O) & [0.1]  & 0.070 \\
Poi & (C) & [sin]  & 0.088 & Poi & (O) & [sin]  & 0.066 \\
Poi & (C) & [pc]   & 0.086 & Poi & (O) & [pc]   & 0.065 \\
\hline
INR & (C) & [0.05] & 0.057 & INR & (O) & [0.05] & 0.033 \\
INR & (C) & [0.1]  & 0.102 & INR & (O) & [0.1]  & 0.065 \\
INR & (C) & [sin]  & 0.111 & INR & (O) & [sin]  & 0.072 \\
INR & (C) & [pc]   & 0.096 & INR & (O) & [pc]   & 0.063 \\
\hline
Cal & (C) & [0.05] & 0.048 & Cal & (O) & [0.05] & 0.028 \\
Cal & (C) & [0.1]  & 0.092 & Cal & (O) & [0.1]  & 0.054 \\
Cal & (C) & [sin]  & 0.099 & Cal & (O) & [sin]  & 0.061 \\
Cal & (C) & [pc]   & 0.096 & Cal & (O) & [pc]   & 0.050 \\
\hline
Pot & (C) & [0.05] & 0.049 & Pot & (O) & [0.05] & 0.030 \\
Pot & (C) & [0.1]  & 0.095 & Pot & (O) & [0.1]  & 0.059 \\
Pot & (C) & [sin]  & 0.102 & Pot & (O) & [sin]  & 0.067 \\
Pot & (C) & [pc]   & 0.089 & Pot & (O) & [pc]   & 0.059 \\
\hline
Nor & (C) & [0.05] & 0.052 & Nor & (O) & [0.05] & 0.023 \\
Nor & (C) & [0.1]  & 0.086 & Nor & (O) & [0.1]  & 0.047 \\
Nor & (C) & [sin]  & 0.100 & Nor & (O) & [sin]  & 0.058 \\
Nor & (C) & [pc]   & 0.098 & Nor & (O) & [pc]   & 0.050 \\
        \bottomrule
    \end{tabular}
\end{table}

\section{Empirical Verification of the Bounds on FDR and FPR}\label{sec:bound:pois}

\begin{figure*}[ht]
    \centering
    \includegraphics[width=0.33\linewidth]{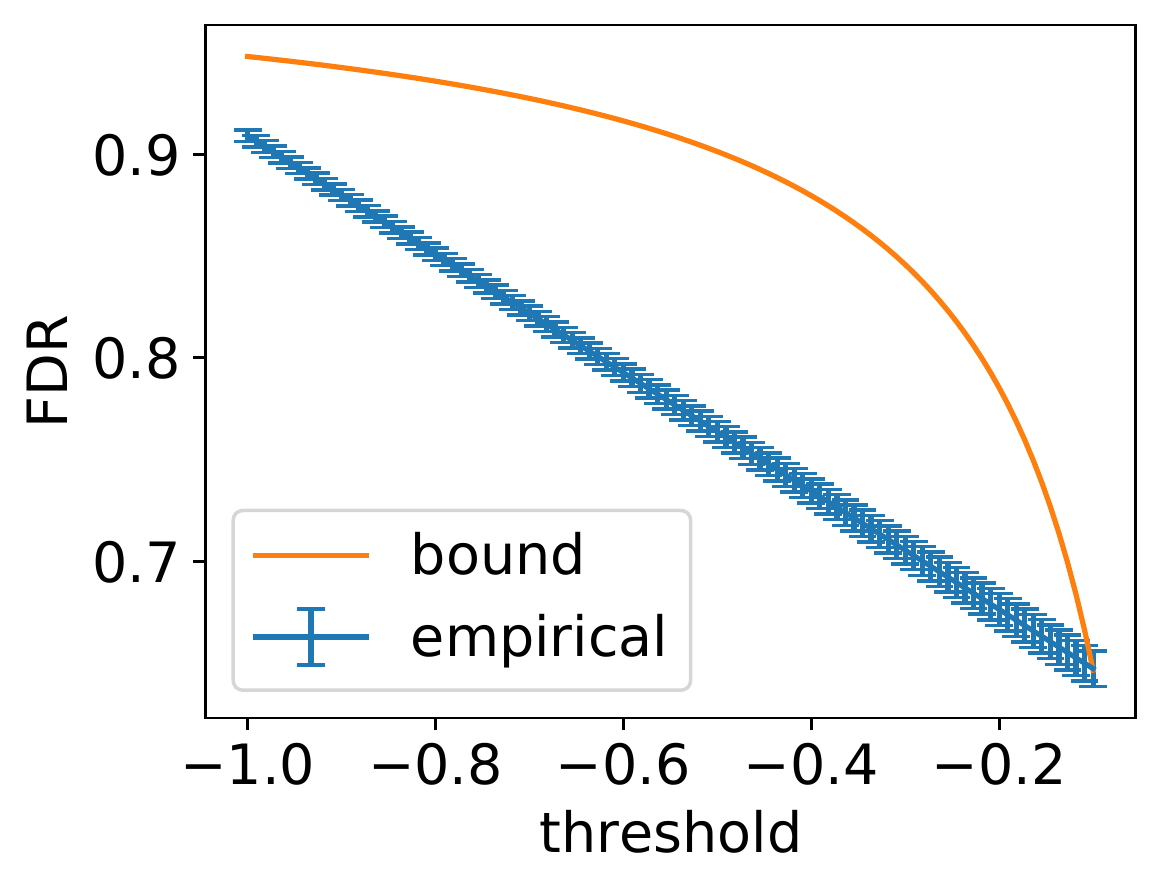}
    \includegraphics[width=0.33\linewidth]{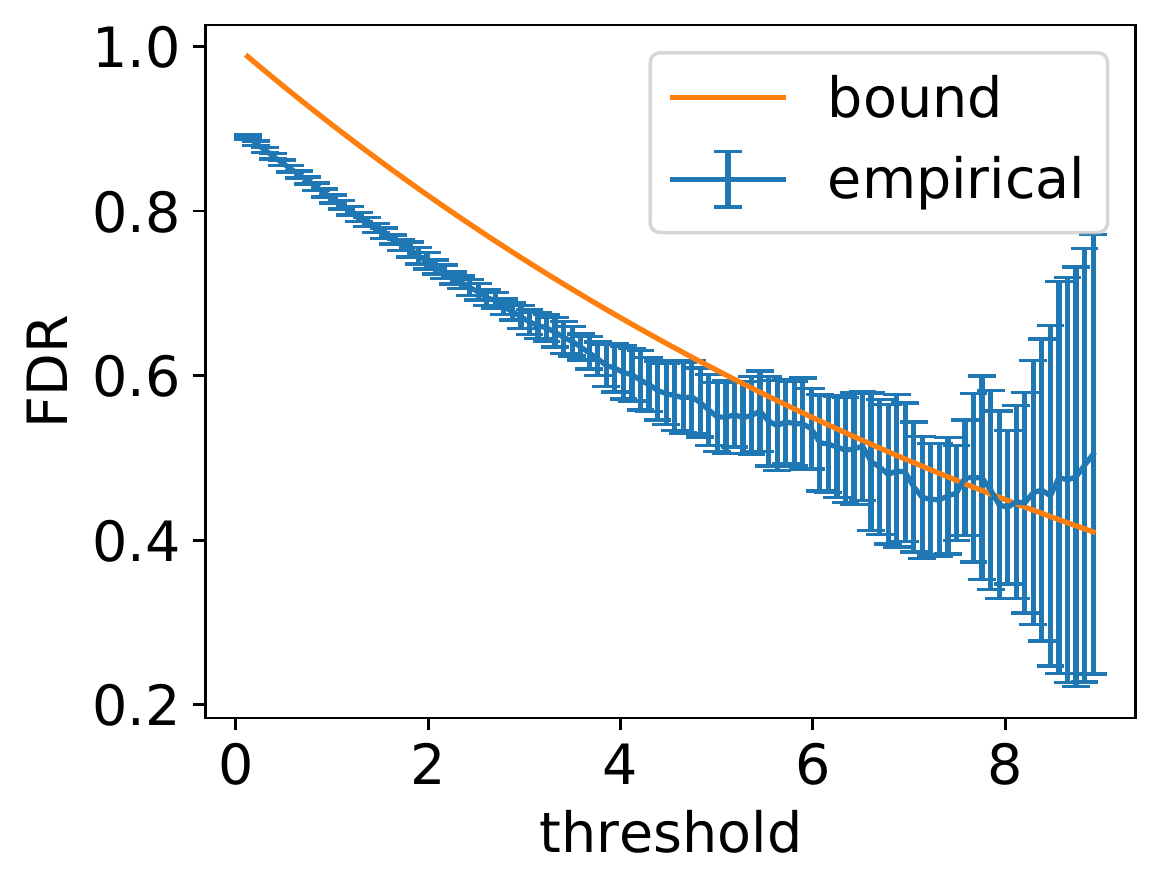}
    \includegraphics[width=0.33\linewidth]{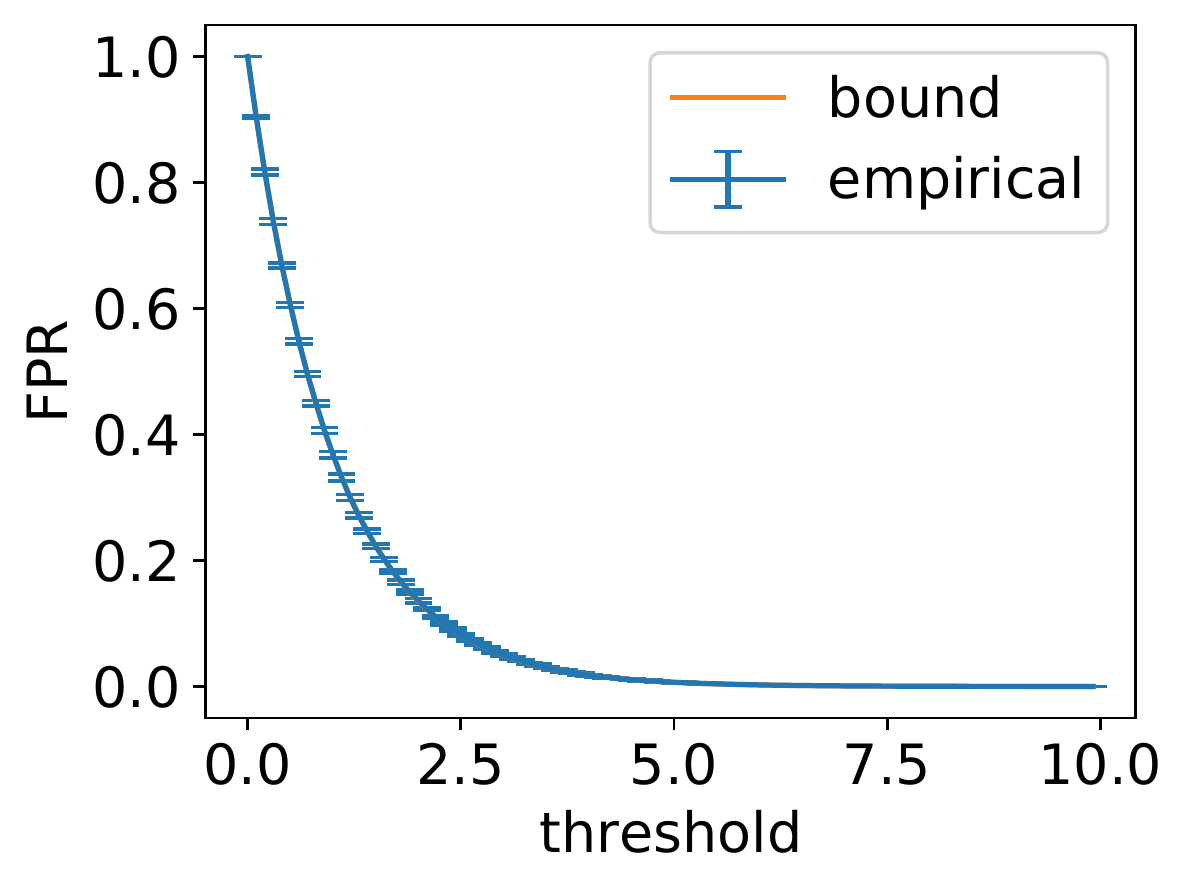}
    \caption{From left to right: FDR (commission outlier), FDR (omission outlier), and FPR (omission outlier) on synthetic data (Poisson process).
    }\label{fig:bound:pois}
\end{figure*}

We show the results of empirically verifying the bounds proved in Section~\ref{sec:bounds}, continuing the results in Section~\ref{sec:exp:synth}. We use  {\True} (Ground Truth): our outlier scoring methods combined with the \emph{ground-truth} point-process model, which is only available on synthetic data. Figure~\ref{fig:bound:pois} shows the empirical FDR (commission outlier), FDR (omission outlier), and FPR (omission outlier) with means and standard deviations on data simulated from inhomogeneous Poisson processes along with the theoretical bounds. As we can see, the empirical FDRs have high variance when the threshold is high, because there are smaller number of samples above a higher threshold. Nonetheless, the empirical FDRs conform with the theoretical bounds, and so does the empirical FPR.

\section{Additional Experiment Results}\label{sec:additional:results}

\paragraph{Using Ground-Truth Model}

\begin{figure*}
    \centering
    \includegraphics[width=0.45\linewidth]{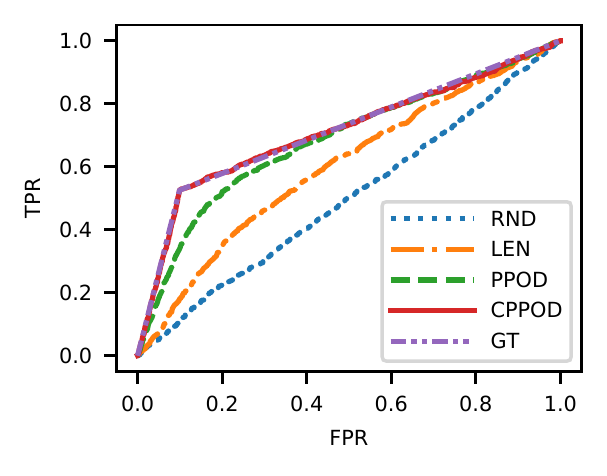}
    \includegraphics[width=0.45\linewidth]{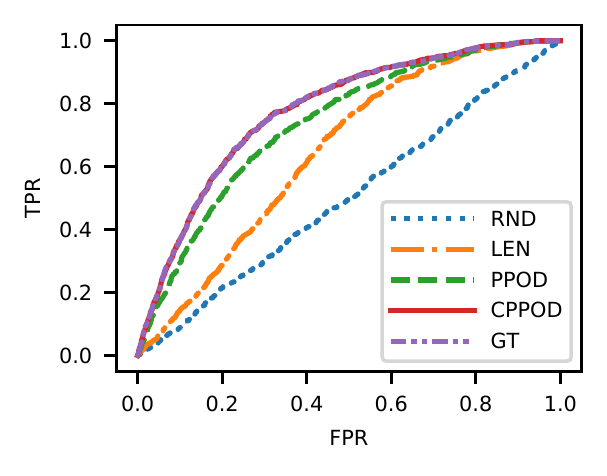}
    \caption{ROC curves on synthetic data (Poisson process). Left: commission. Right: omission.
    }\label{fig:roc:pois}
\end{figure*}

\begin{figure*}
    \centering
    \includegraphics[width=0.45\linewidth]{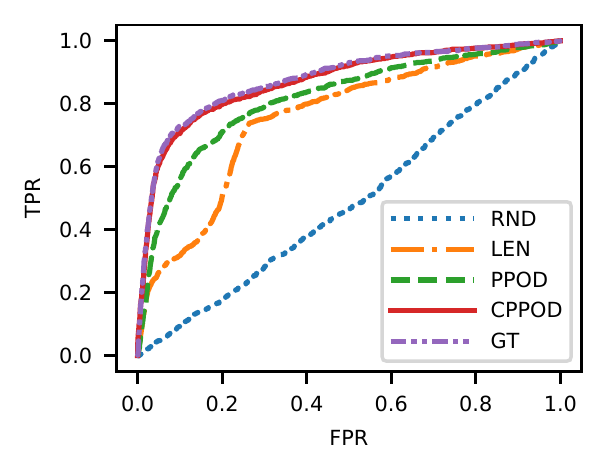}
    \includegraphics[width=0.45\linewidth]{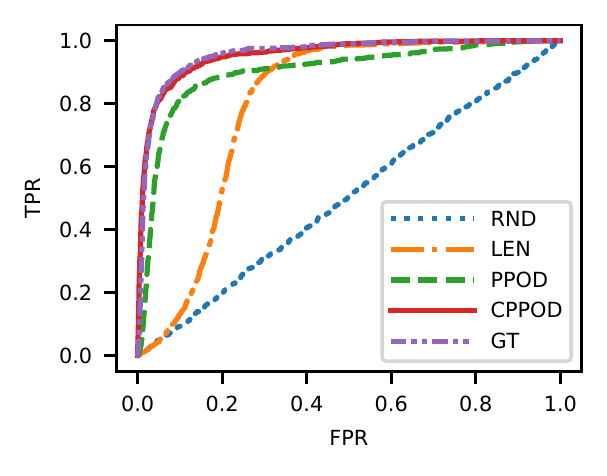}
    \caption{ROC curves on synthetic data (Gamma process). Left: commission. Right: omission.
    }\label{fig:roc:gam}
\end{figure*}

We also compared with {\True} (Ground Truth): our outlier scoring methods combined with the \emph{ground-truth} point-process model (only available on synthetic data). Figure~\ref{fig:roc:pois} and ~\ref{fig:roc:gam} show the receiver operating characteristic (ROC) curves of the outlier detection methods on the synthetic data generated from inhomogeneous Poisson processes and Gamma processes with $\alpha = 0.1$. We note that the curves of {\True} and {\CPPOD} are almost identical. The fact that {\CPPOD} almost has the same performance as {\True} is an evidence that the model based on the continuous-time LSTM is flexible enough to represent these context-dependent point processes. 

\paragraph{Varying Outlier Rate}

We also experimented with changing $\alpha_0 = 0.05$ for the constant rate to see its effect. Table~\ref{tab:roc:synthetic} and Table~\ref{tab:roc:mimic} show the full AUROC results for synthetic and MIMIC data respectively. As we can see, the relative performance for each method does not change in almost all cases.

\begin{table*}
    \centering
    \caption{AUROC on synthetic data. Dataset: name abbreviation (C=commission, O=omission) [$\alpha$].}\label{tab:roc:synthetic}
    \begin{tabular}{lll|llll}
        \toprule
        \multicolumn{3}{c|}{Dataset} &   \Rand &    \Len &              \PPOD &             \CPPOD \\
        \midrule
        Poi & (C) & [0.05] &  0.493 ($\pm$ 0.011) &  0.627 ($\pm$ 0.011) &  0.684 ($\pm$ 0.014) &  \textbf{0.716} ($\pm$ 0.019) \\
Poi & (C) & [0.1]  &  0.500 ($\pm$ 0.010) &  0.601 ($\pm$ 0.008) &  0.684 ($\pm$ 0.010) &  \textbf{0.711} ($\pm$ 0.012) \\
Poi & (C) & [sin]  &  0.493 ($\pm$ 0.007) &  0.575 ($\pm$ 0.006) &  0.661 ($\pm$ 0.016) &  \textbf{0.707} ($\pm$ 0.017) \\
Poi & (C) & [pc]   &  0.512 ($\pm$ 0.009) &  0.584 ($\pm$ 0.011) &  0.664 ($\pm$ 0.009) &  \textbf{0.697} ($\pm$ 0.014) \\
\hline
Poi & (O) & [0.05] &  0.491 ($\pm$ 0.018) &  0.650 ($\pm$ 0.008) &  0.736 ($\pm$ 0.007) &  \textbf{0.776} ($\pm$ 0.009) \\
Poi & (O) & [0.1]  &  0.503 ($\pm$ 0.008) &  0.650 ($\pm$ 0.006) &  0.737 ($\pm$ 0.006) &  \textbf{0.778} ($\pm$ 0.005) \\
Poi & (O) & [sin]  &  0.498 ($\pm$ 0.013) &  0.659 ($\pm$ 0.007) &  0.741 ($\pm$ 0.012) &  \textbf{0.791} ($\pm$ 0.010) \\
Poi & (O) & [pc]   &  0.491 ($\pm$ 0.007) &  0.652 ($\pm$ 0.011) &  0.734 ($\pm$ 0.013) &  \textbf{0.784} ($\pm$ 0.010) \\
\hline
Gam & (C) & [0.05] &  0.479 ($\pm$ 0.018) &  0.776 ($\pm$ 0.011) &  0.840 ($\pm$ 0.010) &  \textbf{0.897} ($\pm$ 0.006) \\
Gam & (C) & [0.1]  &  0.485 ($\pm$ 0.007) &  0.754 ($\pm$ 0.006) &  0.816 ($\pm$ 0.008) &  \textbf{0.871} ($\pm$ 0.006) \\
Gam & (C) & [sin]  &  0.493 ($\pm$ 0.008) &  0.762 ($\pm$ 0.008) &  0.817 ($\pm$ 0.006) &  \textbf{0.886} ($\pm$ 0.004) \\
Gam & (C) & [pc]   &  0.506 ($\pm$ 0.007) &  0.757 ($\pm$ 0.005) &  0.813 ($\pm$ 0.005) &  \textbf{0.870} ($\pm$ 0.007) \\
\hline
Gam & (O) & [0.05] &  0.503 ($\pm$ 0.013) &  0.803 ($\pm$ 0.009) &  0.919 ($\pm$ 0.008) &  \textbf{0.960} ($\pm$ 0.007) \\
Gam & (O) & [0.1]  &  0.505 ($\pm$ 0.012) &  0.799 ($\pm$ 0.005) &  0.901 ($\pm$ 0.007) &  \textbf{0.956} ($\pm$ 0.003) \\
Gam & (O) & [sin]  &  0.503 ($\pm$ 0.010) &  0.809 ($\pm$ 0.006) &  0.902 ($\pm$ 0.006) &  \textbf{0.956} ($\pm$ 0.004) \\
Gam & (O) & [pc]   &  0.515 ($\pm$ 0.010) &  0.813 ($\pm$ 0.005) &  0.905 ($\pm$ 0.006) &  \textbf{0.955} ($\pm$ 0.004) \\

        \bottomrule
    \end{tabular}
\end{table*}

\begin{table*}
    \centering
    \caption{AUROC on MIMIC data. Dataset: name abbreviation (C=commission, O=omission) [$\alpha$].}\label{tab:roc:mimic}
    \begin{tabular}{lll|llll}
        \toprule
        \multicolumn{3}{c|}{Dataset} &   \Rand &    \Len &              \PPOD &             \CPPOD \\
        \midrule
        INR & (C) & [0.05] &  0.486 ($\pm$ 0.014) &  0.613 ($\pm$ 0.018) &  \textbf{0.702} ($\pm$ 0.014) &           0.701 ($\pm$ 0.018) \\
INR & (C) & [0.1]  &  0.496 ($\pm$ 0.010) &  0.596 ($\pm$ 0.009) &           0.682 ($\pm$ 0.010) &  \textbf{0.687} ($\pm$ 0.009) \\
INR & (C) & [sin]  &  0.508 ($\pm$ 0.009) &  0.588 ($\pm$ 0.010) &           0.675 ($\pm$ 0.009) &  \textbf{0.680} ($\pm$ 0.009) \\
INR & (C) & [pc]   &  0.488 ($\pm$ 0.010) &  0.607 ($\pm$ 0.010) &           0.673 ($\pm$ 0.008) &  \textbf{0.681} ($\pm$ 0.010) \\
\hline
INR & (O) & [0.05] &  0.487 ($\pm$ 0.013) &  0.736 ($\pm$ 0.011) &           0.779 ($\pm$ 0.012) &  \textbf{0.782} ($\pm$ 0.012) \\
INR & (O) & [0.1]  &  0.498 ($\pm$ 0.011) &  0.726 ($\pm$ 0.008) &  \textbf{0.748} ($\pm$ 0.009) &           0.746 ($\pm$ 0.010) \\
INR & (O) & [sin]  &  0.516 ($\pm$ 0.012) &  0.717 ($\pm$ 0.011) &           0.760 ($\pm$ 0.010) &  \textbf{0.764} ($\pm$ 0.009) \\
INR & (O) & [pc]   &  0.508 ($\pm$ 0.009) &  0.720 ($\pm$ 0.011) &  \textbf{0.773} ($\pm$ 0.009) &           0.770 ($\pm$ 0.009) \\
\hline
Cal & (C) & [0.05] &  0.470 ($\pm$ 0.020) &  0.753 ($\pm$ 0.017) &           0.843 ($\pm$ 0.012) &  \textbf{0.885} ($\pm$ 0.010) \\
Cal & (C) & [0.1]  &  0.504 ($\pm$ 0.013) &  0.739 ($\pm$ 0.012) &           0.830 ($\pm$ 0.010) &  \textbf{0.866} ($\pm$ 0.006) \\
Cal & (C) & [sin]  &  0.502 ($\pm$ 0.016) &  0.688 ($\pm$ 0.015) &           0.797 ($\pm$ 0.010) &  \textbf{0.835} ($\pm$ 0.009) \\
Cal & (C) & [pc]   &  0.508 ($\pm$ 0.011) &  0.742 ($\pm$ 0.011) &           0.837 ($\pm$ 0.009) &  \textbf{0.860} ($\pm$ 0.011) \\
\hline
Cal & (O) & [0.05] &  0.513 ($\pm$ 0.021) &  0.531 ($\pm$ 0.014) &           0.760 ($\pm$ 0.014) &  \textbf{0.761} ($\pm$ 0.014) \\
Cal & (O) & [0.1]  &  0.493 ($\pm$ 0.016) &  0.526 ($\pm$ 0.009) &           0.759 ($\pm$ 0.008) &  \textbf{0.775} ($\pm$ 0.008) \\
Cal & (O) & [sin]  &  0.518 ($\pm$ 0.017) &  0.529 ($\pm$ 0.012) &           0.758 ($\pm$ 0.009) &  \textbf{0.777} ($\pm$ 0.010) \\
Cal & (O) & [pc]   &  0.496 ($\pm$ 0.017) &  0.541 ($\pm$ 0.010) &           0.759 ($\pm$ 0.011) &  \textbf{0.780} ($\pm$ 0.009) \\
\hline
Pot & (C) & [0.05] &  0.488 ($\pm$ 0.020) &  0.707 ($\pm$ 0.016) &           0.827 ($\pm$ 0.012) &  \textbf{0.878} ($\pm$ 0.009) \\
Pot & (C) & [0.1]  &  0.498 ($\pm$ 0.012) &  0.733 ($\pm$ 0.013) &           0.839 ($\pm$ 0.009) &  \textbf{0.878} ($\pm$ 0.009) \\
Pot & (C) & [sin]  &  0.503 ($\pm$ 0.010) &  0.691 ($\pm$ 0.009) &           0.813 ($\pm$ 0.011) &  \textbf{0.857} ($\pm$ 0.007) \\
Pot & (C) & [pc]   &  0.511 ($\pm$ 0.010) &  0.718 ($\pm$ 0.013) &           0.831 ($\pm$ 0.008) &  \textbf{0.874} ($\pm$ 0.010) \\
\hline
Pot & (O) & [0.05] &  0.503 ($\pm$ 0.015) &  0.539 ($\pm$ 0.014) &           0.727 ($\pm$ 0.015) &  \textbf{0.744} ($\pm$ 0.014) \\
Pot & (O) & [0.1]  &  0.495 ($\pm$ 0.017) &  0.533 ($\pm$ 0.012) &           0.736 ($\pm$ 0.011) &  \textbf{0.748} ($\pm$ 0.011) \\
Pot & (O) & [sin]  &  0.508 ($\pm$ 0.011) &  0.536 ($\pm$ 0.014) &           0.744 ($\pm$ 0.011) &  \textbf{0.759} ($\pm$ 0.012) \\
Pot & (O) & [pc]   &  0.524 ($\pm$ 0.010) &  0.552 ($\pm$ 0.011) &           0.746 ($\pm$ 0.011) &  \textbf{0.761} ($\pm$ 0.011) \\
\hline
Nor & (C) & [0.05] &  0.506 ($\pm$ 0.013) &  0.868 ($\pm$ 0.014) &           0.899 ($\pm$ 0.013) &  \textbf{0.899} ($\pm$ 0.013) \\
Nor & (C) & [0.1]  &  0.494 ($\pm$ 0.014) &  0.864 ($\pm$ 0.010) &           0.890 ($\pm$ 0.012) &  \textbf{0.897} ($\pm$ 0.013) \\
Nor & (C) & [sin]  &  0.536 ($\pm$ 0.012) &  0.837 ($\pm$ 0.012) &           0.858 ($\pm$ 0.014) &  \textbf{0.871} ($\pm$ 0.014) \\
Nor & (C) & [pc]   &  0.524 ($\pm$ 0.012) &  0.844 ($\pm$ 0.016) &  \textbf{0.884} ($\pm$ 0.014) &           0.882 ($\pm$ 0.013) \\
\hline
Nor & (O) & [0.05] &  0.506 ($\pm$ 0.023) &  0.489 ($\pm$ 0.018) &  \textbf{0.829} ($\pm$ 0.013) &           0.826 ($\pm$ 0.012) \\
Nor & (O) & [0.1]  &  0.510 ($\pm$ 0.010) &  0.468 ($\pm$ 0.016) &  \textbf{0.835} ($\pm$ 0.010) &           0.832 ($\pm$ 0.009) \\
Nor & (O) & [sin]  &  0.488 ($\pm$ 0.014) &  0.462 ($\pm$ 0.013) &  \textbf{0.842} ($\pm$ 0.011) &           0.837 ($\pm$ 0.011) \\
Nor & (O) & [pc]   &  0.503 ($\pm$ 0.012) &  0.476 ($\pm$ 0.014) &  \textbf{0.851} ($\pm$ 0.011) &           0.848 ($\pm$ 0.010) \\

        \bottomrule
    \end{tabular}
\end{table*}

\end{document}